\documentclass{article}

\usepackage{hyperref}       % hyperlinks
\usepackage{url}            % simple URL typesetting
\usepackage{booktabs}       % professional-quality tables
\usepackage{amsfonts}       % blackboard math symbols
\usepackage{nicefrac}       % compact symbols for 1/2, etc.
\usepackage{microtype}      % microtypography

\usepackage{graphicx}
\usepackage{subfigure}
\usepackage{amsthm}	
\usepackage{appendix}
\usepackage{latexsym,amsmath,amssymb}

\begin{document}
	
\title{Some Geometrical and Topological Properties of DNNs' Decision Boundaries}	
\author{Bo Liu,  Mengya Shen \\
	College of Computer Science, Faculty of Information Technology\\
	Beijing University of Technology, Beijing, China \\
	\texttt{liubo@bjut.edu.cn, Shenmy@emails.bjut.edu.cn}}
\date{}
\maketitle

\begin{abstract}
Geometry and topology of decision regions are closely related with classification performance and robustness against adversarial attacks. In this paper, we use differential geometry to theoretically explore the geometrical and topological properties of decision regions produced by deep neural networks (DNNs). The goal is to obtain some geometrical and topological properties of decision boundaries for given DNN models, and provide some principled guidance to design and regularization of DNNs. First, we present the curvatures of decision boundaries in terms of network parameters, and give sufficient conditions on network parameters for producing flat or developable decision boundaries. Based on the Gauss-Bonnet-Chern theorem in differential geometry, we then propose a method to compute the Euler characteristics of compact decision boundaries, and verify it with experiments.
\end{abstract}

\section{Introduction}

Although deep learning has been successfully applied to various domains including computer vision, speech recognition and natural language processing, the theoretical understanding of its success is still limited. One important theoretical aspect of deep learning is the geometry and topology of decision regions produced by DNNs. Geometry of decision boundaries concerns their curvature, size and convexity etc., and topology of decision boundaries considers their connectivity and genus (the number of holes) etc. There are two important relevant questions: on one hand, how to obtain the geometrical and topological properties of decision boundaries for given DNN models? And on the other hand, what are the constraints on network architecture and parameters in order to produce decision boundaries with desired geometrical and topological features? Besides theoretical interest, answering these questions is important to model selection \cite{Guss18,Ramamurthy_tda_icml19}, reducing overfitting \cite{Chen19topo} and improving DNNs' robustness against adversarial attacks \cite{Fawzi19,Fawzi18}. Moreover, researches on the geometry and topology of decision regions can provide some principled guidance for network design. For example, choose appropriate network architecture to produce disconnected decision boundaries \cite{Nguyen18_disconnected}, and regularize curvature or number of connected components to boost robustness or accuracy \cite{Fawzi19,Chen19topo,Hofer_icml19}. 

Researches on the geometry and topology of DNN produced decision regions are still at their infancy. This paper aims at exploring some geometrical and topological properties of DNNs' decision regions using knowledges from differential geometry and topology \cite{DoCarmo,differentialgeometry}. We first use differential geometry to compute the curvature of decision boundaries. We formulate curvatures in terms of network parameters, and based on that it is possible to design DNNs with desired geometrical properties. We then use global differential geometry, which establishes the analytical connection between local curvature and global topology of manifolds, to compute some topological properties of decision boundaries.

The contributions of this paper are summarized as follows:
\begin{itemize}
	\item
	For DNNs with any input dimension, we give sufficient conditions on network parameters for producing flat or developable decision boundaries.
	\item
	Based on global differential geometry, more specifically, the Gauss-Bonnet-Chern theorem for 2D surfaces and higher-dimensional manifolds, we propose a method to compute the Euler characteristics (a topological property) of compact decision boundary manifolds.
	
\end{itemize}	

This paper is organized as follows. Section \ref{section1.1} is related work. In section \ref{section2}, we present the preliminaries of differential geometry, and introduce the feed-forward deep neural network models, as well as some definitions and notations. Curvatures of decision boundaries for 2D, 3D and higher-dimensional input spaces are explored in section \ref{section3}. Section \ref{section6} presents the sufficient conditions on network parameters for producing flat or developable decision boundaries. In section \ref{section4}, we propose the method to compute Euler characteristics of decision boundaries, with experiments on synthetic data. The proofs of main results are given in section \ref{section7}. Finally, we give our conclusion and future work.

\subsection{Related work}\label{section1.1}
There have been some works on the geometry and topology of DNN produced decision regions. \cite{Poole16} experimentally finds that the decision boundary can become exponentially curved with depth, thus enabling highly complex classifications. \cite{Fawzi19,Fawzi18} show that decision boundaries exhibiting quasi-linear behavior in the vicinity of data points can improve adversarial robustness. Although numerical computations of curvatures have been presented in these works, systematic expressions of curvatures in terms of network parameters are still missing. \cite{Fawzi18} demonstrates that the decision regions of CaffeNet \cite{Jia_caffe} trained on ImageNet are connected. Based on persistent homology \cite{Persistent_homology}, \cite{Ramamurthy_tda_icml19,Guss18} compute the topological properties of DNNs’ decision boundaries by constructing simplicial complexes of decision boundaries with discrete data points. In comparison, our method is analytical and based on global differential geometry. \cite{Nguyen18_disconnected} demonstrates that a sufficiently wide hidden layer is necessary to produce disconnected decision regions. \cite{Beise19} shows that the decision regions are unbounded for DNNs with width less than or equal to input dimensions. \cite{Bianchini} shows the topological expressiveness advantage of deep networks over shallow ones in terms of bounds on the sum of Betti numbers. To achieve a better balance between accuracy and overfitting, \cite{Chen19topo,Hofer_icml19} propose persistent homology based connectivity regularizers for logistic regression classifier and autoencoder respectively.
\cite{BARCODES_iclr20} presents an algorithm to compute the topological features of neural networks’ objective functions.

\section{Preliminaries}\label{section2}	

\subsection{Fully-connected feed-forward deep neural networks}

We consider in this paper fully-connected feed-forward deep neural networks for binary classification. Let $ d $ be the input dimension, $ L $ be the number of hidden layers and $ d_i $ be the width of layer $ i $ with $ d_0 = d $. The output of a fully-connected feed-forward DNN is 
\begin{equation}\label{eq50}
\begin{aligned}
f(\mathbf{x})=\mathbf{a}^{T} \sigma ( \mathrm{W}^{L} \sigma\left(\mathrm{W}^{L-1} \cdots \sigma\left(\mathrm{W}^{1} \mathbf{x}+\mathbf{b}^{1}\right)+\cdots+\mathbf{b}^{L-1}\right) +\mathbf{b}^{L} ) +c,
\end{aligned}
\end{equation}
where $\mathbf{x} \in \mathbb{R}^{d}$ is the input, $\mathrm{W}^{i} \in \mathbb{R}^{d_{i} \times d_{i-1}}$ and $\mathbf{b}^{i} \in \mathbb{R}^{d_{i}}$ are the weight matrix and bias
vector of layer $ i $ respectively, $\mathbf{a} \in \mathbb{R}^{d_{L}}$ and $c$ are the weight vector and bias of the output layer. $\sigma: \mathbb{R} \rightarrow \mathbb{R}$ is the activation function of every hidden layer which applies component-wise. We consider in this paper activation functions that are at minimum twice differentiable and strictly monotonically increasing (thus bijective), including the widely used sigmoid, tanh, and softplus ( $\sigma(t)=\frac{1}{\alpha} \log (1+
e^{\alpha t})$ ) which can approximate ReLU activation arbitrarily well.

We consider binary classification problems in this paper. The solution to $f(\mathbf{x})=0$ gives the decision boundary, $\{\mathbf{x} \in \mathbb{R}^{d} | f(\mathbf{x})>0 \}$ gives the decision region of the positive class and $\{\mathbf{x} \in \mathbb{R}^{d} | f(\mathbf{x})<0 \}$ specifies the decision region of the negative class. We assume that the decision boundaries are smooth manifolds. 

The pre-image of a mapping $f: U \rightarrow V$ is defined as the set $\{x \in U | f(x) \in V\} .$ $ [d] $ stands for
$\{1,2, \cdots, d\}.$ Both $\mathbf{w}_{i}^T$ and $\mathrm{W}_{i,\cdot }$ represent the $i$th row of matrix W, and $\mathrm{W}_{\cdot, i}$ denotes the $ i $th column of W.

\subsection{Differential geometry and topology}
In this subsection, we will give a brief introduction to the concepts and knowledges of differential geometry and topology that will be used in this paper. Interested readers are referred to \cite{DoCarmo,differentialgeometry} for the details.

\subsubsection{Differential geometry of curves}
A planar curve is locally parameterized as $\mathbf{r}(t)=\left(\begin{array}{l}x(t) \\ y(t)\end{array}\right),$ which maps the parameter $t \in \mathbb{R}$ to a point $\mathbf{r}(t) \in \mathbb{R}^{2}$ on the curve. Arc length $s$ is usually used as the parameter $t$. The tangent vector at a point on the curve is defined as $\mathbf{T}=\frac{\mathrm{d}\mathbf{r}(s)}{\mathrm{ds}}.$ The Frenet formula $\frac{d\mathbf{T}(s)}{\mathrm{d}s}=k \mathbf{N}$
gives the definition of planar curvature $k$, where $\mathbf{N}$ is the unit normal vector.

\subsubsection{Differential geometry of surfaces}

A 2D surface is locally parameterized as $\mathbf{r}(u, v)=(x(u, v), y(u, v), z(u, v))^{T},$ which maps the
parameters $(u, v) \in \mathbb{R}^{2}$ to a point $\mathbf{r}(u, v) \in \mathbb{R}^{3}$ on the surface. Let d$ s $ be the infinitesimal arc length on the surface, then
$ \mathrm{d}s^{2}=\mathrm{d} \mathbf{r} \cdot \mathrm{d} \mathbf{r}=E \mathrm{d} u^{2}+2 F \mathrm{d} u \mathrm{d} v+G \mathrm{d} v^{2}
$, where $E=\mathbf{r}_{u} \cdot \mathbf{r}_{u}, F=\mathbf{r}_{u} \cdot \mathbf{r}_{v}, G=\mathbf{r}_{v} \cdot \mathbf{r}_{v}$ are called coefficients of the first fundamental
form. $\mathbf{r}_{u}$ denotes $\frac{\partial \mathbf{r}(u, v)}{\partial u}$ and $\mathbf{r}_{v}$ denotes $\frac{\partial \mathbf{r}(u, v)}{\partial v}$. The second fundamental form, which measures the distance from $\mathbf{r}(u+\mathrm{d} u, v+\mathrm{d} v)$ to the tangent plane at $\mathbf{r}(u, v),$ is given as $L \mathrm{d} u^{2}+2 M \mathrm{d} u \mathrm{d} v+N \mathrm{d} v^{2},$ where $L=\mathbf{r}_{u u} \cdot \mathbf{n}, N=$
$\mathbf{r}_{v v} \cdot \mathbf{n}, M=\mathbf{r}_{u v} \cdot \mathbf{n}$ are called coefficients of the second fundamental form. $\mathbf{n}:=\frac{\mathbf{r}_{u} \times \mathbf{r}_{v}}{\left|\mathbf{r}_{u} \times \mathbf{r}_{v}\right|}$ is
the unit normal vector, $\mathbf{r}_{uu} := \frac{\partial^2 \mathbf{r}(u, v)}{\partial u^2}$, $\mathbf{r}_{uv} := \frac{\partial^2 \mathbf{r}(u, v)}{\partial u \partial v}$ and $\mathbf{r}_{vv} := \frac{\partial^2 \mathbf{r}(u, v)}{\partial v^2}$.

At any point on the surface, there are two principal curvatures $k_{1}$ and $k_{2}$ defined along two orthogonal principal directions. The Gaussian curvature is defined as
$K=k_{1} k_{2}=\frac{L N-M^{2}}{E G-F^{2}}$.

The classification theorem of compact surfaces states that all orientable 2D compact surfaces can be classified into homeomorphism classes by their Euler characteristics. The Euler characteristic $\chi(S)$ of a 2D compact surface $S$ can take a value only in $\{2,0,-2, \ldots,-2 n, \ldots\} .$ For
example, $\chi=2$ for a sphere, $\chi=0$ for a torus and $\chi=-2$ for a double torus. $n=\frac{2-\chi(S)}{2}$ is
called the genus of $S$, which is the number of holes in $S$.

For an orientable 2D compact surface $S$, the Gauss-Bonnet theorem states $\iint_{S} K \mathrm{d} \sigma=$ $2 \pi \chi(S),$ where $\mathrm{d} \sigma$ is the area element. Gauss-Bonnet theorem establishes the connection between local curvature and global topological property.

\subsubsection{Differential geometry of higher-dimensional manifolds}

High-dimensional case is much complex and unintuitive. A $ n $-dimensional manifold is locally parameterized as $\mathbf{r}=\mathbf{r}\left(x_{1}, x_{2}, \ldots, x_{n}\right).$ Then
$$
\mathrm{d}s^{2}=\mathrm{d} \mathbf{r} \cdot \mathrm{d} \mathbf{r}=\sum_{i, j=1}^{n} g_{i j} \mathrm{d} x_{i} \mathrm{d} x_{j}
$$
where $g_{i j}=\mathbf{r}_{i} \cdot \mathbf{r}_{j}$ and $ \mathbf{r}_{i}= \frac{\partial \mathbf{r}}{\partial x_i} $.  $g=\left(g_{i j}\right)$ is called metric tensor. The Riemannian connection is defined as
$\Gamma_{i k}^{m} \ := \frac{1}{2} \Sigma_{j} g^{m j}\left(\partial_{k} g_{i j}+\partial_{i} g_{k j}-\partial_{j} g_{i k}\right),$ where $g^{m j}$ is the element of $g^{-1},$ the inverse matrix
of $g$. $ \partial_k g_{ij} = \frac{\partial g_{ij}}{\partial x_k} $.

The Riemann curvature tensor is defined as
$$
R_{b i k}^{a} :=\Gamma_{k b, i}^{a}-\Gamma_{i b, k}^{a}+\sum_{c} \Gamma_{i c}^{a} \Gamma_{k b}^{c}-\sum_{c} \Gamma_{k c}^{a} \Gamma_{i b}^{c}
$$
At any point on the manifold, the curvature along any 2D tangent subspace (like $K$ for 2D surfaces) can be obtained from $R_{b i k}^{a} .$ The curvature 2-form is defined as
$$
\Omega_{a b}=\frac{1}{2} \sum_{i, k} R_{a b i k} \mathrm{d} x^{i} \wedge \mathrm{d} x^{k}=\frac{1}{2} \sum_{i, k, c} g_{a c} R_{b i k}^{c} \mathrm{d} x^{i} \wedge \mathrm{d} x^{k},
$$
where $\wedge$ is the outer product of differential forms.

The Euler characteristic $\chi(M)$ of a manifold $M$ is defined as the signed sum of its Betti numbers, which in turn is a concept from algebraic topology. Intuitively, the $ 0 $-st Betti number $ b_0 $ is the number of connected components, and the $i$th Betti number $ b_i\ (i\geq 1) $ counts the number of $i$-dimensional holes of the manifold.

The Gauss-Bonnet-Chern theorem for high-dimensional orientable compact manifolds gives the relationship between local curvature and global topological property. For a $2n$-dimensional orientable compact manifold $M,$ Gauss-Bonnet-Chern theorem states
$$
\chi(M)=\int_{M} e(M),
$$
where $e(M)=\frac{1}{2^{n}(2 \pi)^{n} n !} \sum_{a_{1}, a_{2}, \cdots, a_{2 n}} \varepsilon_{a_{1}, a_{2}, \cdots, a_{2 n}} \Omega^{a_{1} a_{2}} \wedge \cdots \wedge \Omega^{a_{2 n-1} a_{2 n}}, \ \Omega^{a b}=\Sigma_{c, d} g^{a c} g^{b d} \Omega_{c d}, \ $
$\varepsilon_{a_{1} \cdots, a_{2 n}}=\sqrt{\operatorname{det}(g)} \delta_{a_{1} \cdots a_{2 n}}^{1 \cdots 2 n}.\\ 
\delta_{a_{1} \cdots a_{2 n}}^{1 \cdots 2 n}$ is the generalized Kronecker symbol which equals 1 if $(a_{1} \ \cdots \ a_{2 n})$ is an even permutation of $(1 \ 2 \ \cdots \ 2n )$, $-1$ if odd permutation and 0 otherwise.

\section{Curvature of decision boundary}\label{section3}
While there are numerical computations of curvatures in (e.g. \cite{Poole16,Fawzi19}), a systematic and analytical treatment of decision boundary's curvature in the framework of differential geometry is still absent. In this section, we will compute the curvature of decision boundaries for 2D, 3D and higher-dimensional input spaces respectively. The key observation is that although  the decision boundaries usually cannot be solved analytically, we can use the equation $f(\mathbf{x})=0$ to describe them implicitly and use the derivatives of implicit functions to compute curvatures. The expressions of curvatures obtained in this section will be used frequently later in this paper. 

\subsection{2D input space}
In 2D input space, the decision boundary is in general a curve and satisfy $f(x, y)=0$. This gives an implicit function $y=y(x)$. 

\newtheorem{thm}{Theorem}[section]
\newtheorem{lemma}{Lemma}[section]
\begin{lemma}\label{lemma3.1} 
	The planar curvature of a decision boundary in 2D input space is
	\begin{equation}\label{eq2}
	k=| \frac{f_{x x} f_{y}^{2}-2 f_{x} f_{y} f_{y x}+f_{x}^{2} f_{y y}}{f_{y}^{3}} | \left(1+\frac{f_{x}^{2}}{f_{y}^{2}}\right)^{-\frac{3}{2}},
	\end{equation}
	where $f=f(x, y)$ is the output function of neural network, $f_{x}=\frac{\partial f}{\partial x}$, $ f_{xx}=\frac{\partial^2 f}{\partial x^2} $ and so on.
\end{lemma}

For neural networks, we can compute the derivatives involved in \eqref{eq2} and consequently $k$ in terms of network weights. For a one-hidden-layer network $f(\mathbf{x})=\mathbf{a}^{\mathrm{T}} \sigma(\mathrm{W}\mathbf{x}+\mathbf{b})+c,$ its $1st$-order and $2nd$-order partial derivatives are as follows,
\begin{equation}\label{eq3}
\begin{array}{c}
f_{x}=\sum_{i} \mathrm{a}_{i} \sigma_{i}^{\prime} \mathrm{W}_{i 1}, \ \ \ f_{y}=\sum_{i} \mathrm{a}_{i} \sigma_{i}^{\prime} \mathrm{W}_{i 2} \\
f_{x x}=\sum_{i} \mathrm{a}_{i} \sigma_{i}^{\prime \prime} \mathrm{W}_{i 1} \mathrm{W}_{i 1}, \ \ \ f_{y x}=\sum_{i} \mathrm{a}_{i} \sigma_{i}^{\prime \prime} \mathrm{W}_{i 1} \mathrm{W}_{i 2} \\
f_{x y}=\sum_{i} \mathrm{a}_{i} \sigma_{i}^{\prime \prime} \mathrm{W}_{i 1} \mathrm{W}_{i 2}, \ \ \ f_{y y}=\sum_{i} \mathrm{a}_{i} \sigma_{i}^{\prime \prime} \mathrm{W}_{i 2} \mathrm{W}_{i 2},
\end{array}
\end{equation}
where $\sigma_{i}^{\prime}=\sigma^{\prime}\left(\mathbf{w}_{i} \cdot \mathbf{x}+\mathrm{b}_{i}\right), \sigma_{i}^{\prime \prime}=\sigma^{\prime \prime}\left(\mathbf{w}_{i} \cdot \mathbf{x}+\mathrm{b}_{i}\right)$.

For general fully-connected deep neural networks in \eqref{eq50}, define
\begin{equation*}
\mathbf{x}^l =  \mathrm{W}^{l} \sigma\left(\mathrm{W}^{l-1} \cdots \sigma\left(\mathrm{W}^{1} \mathbf{x}+\mathbf{b}^{1}\right)  \cdots+\mathbf{b}^{l-1}\right) +\mathbf{b}^{l} \in \mathbb{R}^{d_l} \ (1 \leq l \leq L) ,
\end{equation*}
then the 1st-order derivative is as follows
\begin{equation}\label{eq51}
f_{x_i} = \mathbf{a}^{T} J(\mathbf{x}^L) \mathrm{W}^{L} J(\mathbf{x}^{L-1}) \mathrm{W}^{L-1} \cdots J(\mathbf{x}^1) \mathrm{W}^{1}_{.,i} , \ (i=1,2) ,
\end{equation}
where $ x_1:=x, x_2:=y $, $ J(\mathbf{x}^l) := diag( \sigma^{\prime}(x^l_1) , \sigma^{\prime}(x^l_2), \cdots, \sigma^{\prime}(x^l_{d_l}) )$.
The 2nd-order derivatives are tedious and omitted here to save space.

\subsection{3D input space}
In 3D input space, the decision boundary is a 2D surface satisfying $f(x, y, z)=0$. Its Gaussian curvature is described by the following lemma.
\begin{lemma}\label{lemma3.2}
	The Gaussian curvature of a decision boundary in 3D input space is
	\begin{equation}\label{eq4}  
	\begin{aligned}
	K=[ &\left(2 f_{x} f_{z} f_{x z}-f_{x x} f_{z}^{2}-f_{z z} f_{x}^{2}\right)\left(2 f_{y} f_{z} f_{y z}-f_{y y} f_{z}^{2}-f_{z z} f_{y}^{2}\right)- \\ 
	&\left(f_{x} f_{z} f_{y z}+f_{y} f_{z} f_{x z}-f_{x} f_{y} f_{z z}-f_{x y} f_{z}^{2}\right)^{2} ] / {[f_{z}^{2}\left(f_{x}^{2}+f_{y}^{2}+f_{z}^{2}\right)^{2}]} .
	\end{aligned}
	%\begin{array}{l}
	%\end{array}
	\end{equation}
\end{lemma}

\subsection{Higher-dimensional input space}
For high-dimensional case, we need to compute the curvature tensor and curvature 2-form. Before that, it is necessary to compute metric tensor $(g_{i j})$ and Riemannian connection $\Gamma_{i k}^{m}$. The details of computation will be presented in subsection \ref{section7.3}.

\section{Conditions on network parameters for producing flat or developable decision boundaries}\label{section6}
Given the expressions of curvatures of decision boundaries, one way wonder what network parameters can generate decision boundaries with some desired geometrical properties, such as flatness and convexity etc. As a starting point, we discuss in this paper how to get flat or developable decision boundaries  with appropriate network weights. It has been shown in \cite{Fawzi19,Fawzi18} that there is a strong relation between small curvature decision boundaries and large robustness against adversarial attacks.

We will start with low-dimensional input space and networks with one hidden layer, and then extend to the general case of arbitrary input dimensions and DNNs. We will derive the sufficient conditions for flat or developable decision boundaries, and give geometric interpretations of some conditions to show intuitively how they are produced. 

\subsection{2D input space and one-hidden-layer networks}\label{section6.1}
In 2D input space, the flat decision boundary is a line. The sufficient conditions for 2D input space and one-hidden-layer networks are given by the following theorem.
\begin{thm}\label{theorm6.1}
	For one-hidden-layer neural networks $f(\mathbf{x})=\mathbf{a}^{T} \sigma(\mathrm{W}\mathbf{x}+\mathbf{b})+c \ \left(\mathbf{a} \in \mathbb{R}^{d_{1}}\right)$
	with 2D input $\mathbf{x}$ and twice differentiable and bijective activation function $\sigma,$ one of the following conditions is sufficient to produce linear decision boundaries with zero curvatures: 
	$\\$
	$ a) $ $\mathrm{a}_{i} \mathrm{W}_{i 1}=0$ or $\mathrm{a}_{i} \mathrm{W}_{i 2}=0 \ (\forall i \in
	\left[d_{1}\right])$. $\\$
	$ b) $  $\mathrm{a}_{i} \mathrm{a}_{j}\left(\mathrm{W}_{i 1} \mathrm{W}_{j 2}-\mathrm{W}_{i 2} \mathrm{W}_{j 1}\right)=0  \ (\forall i, j \in \left[d_{1}\right])$.
\end{thm}

$\mathrm{a}_{i} \mathrm{W}_{i 1}=0$ or $\mathrm{a}_{i} \mathrm{W}_{i 2}=0$ means $\mathrm{W}_{i 1}=0$ or $\mathrm{W}_{i 2}=0 $ if $\mathrm{a}_{i} \neq 0$. Consequently, if $\forall i, \mathrm{a}_{i} \neq$
$0,$ we have $\mathrm{W}_{\cdot, 1}=0$ or $\mathrm{W}_{\cdot, 2}=0 .$ Geometrically, $\mathrm{W}_{\cdot, 1}=0$ or $\mathrm{W}_{\cdot,2}=0$ transforms the 2D input space into a line which, after activation by $\sigma,$ results in a curve. This curve will intersect the (hyper)plane (or line) in $\mathbb{R}^{d_{1}}$ specified by $ (\mathbf{a}, c) $ at a point (the case in which the curve lies on the (hyper)plane is omitted, since it implies that $ f(\mathbf{x})=0 \ (\forall \mathbf{x} \in \mathbb{R}^2) $ and thus the whole input space is decision boundary), whose pre-image in input space is a line. Thus, $\mathrm{W}_{\cdot, 1}=0$ or $\mathrm{W}_{\cdot,2}=0$ produces linear decision boundaries. Fig.2(a) illustrates the case of $\mathrm{W}_{\cdot,1}=0$.

If $\mathrm{a}_{i} \neq 0$ and $\mathrm{W}_{i 1}=0$ or $\mathrm{W}_{i 2}=0$ only for partial values of $i,$ for example, if $d_{1}=3$ and $\mathbf{a}=$ $(0,1,1)^{T}, \mathrm{W}_{21}=\mathrm{W}_{31}=0,$ then the image of $\mathrm{W}\mathbf{x}+\mathbf{b}=\left(\begin{array}{c}\mathrm{W}_{11} x+\mathrm{W}_{12} y \\ \mathrm{W}_{22} y \\ \mathrm{W}_{32} y\end{array}\right)+\mathbf{b}$ is a plane $p$
with normal $\boldsymbol{n}=\left(0, \mathrm{W}_{32},-\mathrm{W}_{22}\right)^{T}$, and $\sigma(p)$ becomes a curved surface whose normal's $x$ component is also $0 .$ The intersection of $\sigma(p)$ with the plane $ q $ in $\mathbb{R}^{d_{1}}$ with normal $ \mathbf{a} $ is a straight line parallel to $x$ axis, whose pre-image in input space is a line parallel to $x$ axis. This case is illustrated in fig.$2(b)$.

The case of $\left(\mathrm{W}_{i 1} \mathrm{W}_{j 2}-\mathrm{W}_{i 2} \mathrm{W}_{j 1}\right)=0 \ \left(\forall i, j \in\left[d_{1}\right]\right)$ means $\mathrm{W}$ is rank-deficient and the image of $\mathrm{W}\mathbf{x}+\mathbf{b} \ (\mathbf{x} \in \mathbb{R}^{2})$ is a line, similar to the case shown in fig. $2(a)$.

\subsection{3D input space and one-hidden-layer networks}	
For 3D input space and one-hidden-layer networks, the sufficient conditions for flat or developable decision boundaries are presented in the following theorem.
\begin{thm}\label{theorm6.3}	
	For one-hidden-layer networks $f(\mathbf{x})=\mathbf{a}^{T} \sigma(\mathrm{W}\mathbf{x}+\mathbf{b})+c \ \left(\mathrm{W} \in \mathbb{R}^{d_{1} \times 3}\right)$ with
	3D input and twice differentiable and bijective activation function $\sigma$, one of the following conditions is sufficient to produce decision boundaries with zero Gaussian curvatures: 
	$  \\ $ 
	$ a) $ $\mathrm{a}_{i} \mathrm{W}_{i 1}=0$ or $\mathrm{a}_{i} \mathrm{W}_{i 2}=0$ or $\mathrm{a}_{i} \mathrm{W}_{i 3}=0  \ (\forall i \in \left[d_{1}\right]).$   $ \\ $ 
	$ b) $ $\mathrm{a}_{i} \mathrm{a}_{j}\left(\mathrm{W}_{i 2} \mathrm{W}_{j 3}-\mathrm{W}_{i 3} \mathrm{W}_{j 2}\right)=0  \ (\forall i, j \in \left[d_{1}\right])$.
\end{thm}

Fig. $2(\mathrm{c})$ illustrates the case of $\mathrm{W}_{\cdot,1}=0$ for $d_{1}=3$. The 3D input space is mapped into a plane $p$. Denote by $q$ the plane in $\mathbb{R}^{d_{1}}$ determined by $ (\mathbf{a}, c) $, $\sigma^{-1}(q)$ will be a curved surface. The intersection of $p$ with $\sigma^{-1}(q)$ is a curve, whose pre-image in input space is a developable surface that is curved along one direction and straight along the orthogonal direction (using the terminology of differential geometry, one of the principal curvatures $k_{1}$ and $k_{2}$ is zero). It is worthy to point out that in 3D input space, $K=k_{1} k_{2}=0$ does not always imply planar decision boundaries ($k_{1}=k_{2}=0$), developable surfaces (one of $k_{1}$ and $k_{2}$ is zero) are also possible, as shown in fig. $2(\mathrm{c})$.

\subsection{General case: higher dimensional input space and deep neural networks}
From previous subsections, it can be seen that for one-hidden-layer networks, while deriving from different expressions of curvatures for different input dimensions, there are some common conditions for flat or developable decision boundaries. By extension, we will give in this subsection the sufficient conditions for producing flat or developable decision boundaries with DNNs of arbitrary input dimensions. Instead of starting from complex curvature tensor, our proof (in subsection \ref{section7.6}) will be based on analyzing the equation $f(\mathbf{x})=0$. We shall point out that although flat decision boundaries can be obtained without utilizing the expressions of curvatures, many other geometry and topology related tasks, such as those in section \ref{section4}, still rely heavily on curvatures.

\begin{thm}\label{theorm6.4}	
	For fully-connected feed-forward deep neural networks $f(\mathbf{x})=\mathbf{a}^{T} \sigma\left(\mathrm{W}^{L} \sigma\left(\mathrm{W}^{L-1} \ldots \sigma\left(\mathrm{W}^{1} \mathbf{x}+\mathbf{b}^{1}\right)+\cdots+\mathbf{b}^{L-1}\right)+\mathbf{b}^{L}\right)+c,$ the following conditions
	are sufficient to produce linear decision boundaries: $a_{i} \mathrm{W}_{i j}^{L} \mathrm{W}_{j k}^{L-1} \ldots \mathrm{W}_{m n}^{2}=0 \  (\forall i \in \mathbb{R}^{d_{L}}, j \in \mathbb{R}^{d_{L-1}}, k \in \mathbb{R}^{d_{L-2}}, \cdots, m \in \mathbb{R}^{d_{2}} )  $ and vectors $\mathrm{W}_{l,\cdot}^{1} \ (\forall l \neq n)$ are linear dependent. The following conditions are sufficient to produce decision boundaries that are straight along axis $x_{q}$ :
	$\mathrm{a}_{i} \mathrm{W}_{i j}^{L} \mathrm{W}_{j k}^{L-1} \cdots \mathrm{W}_{m n}^{2}=0 \ (\forall i \in \mathbb{R}^{d_{L}}, j \in \mathbb{R}^{d_{L-1}}, k \in \mathbb{R}^{d_{L-2}}, \cdots, m \in \mathbb{R}^{d_{2}} )$ and $\mathrm{W}_{l q}^{1}=0 \ (\forall l \neq n)$.
\end{thm}

\textbf{Remark:} We have obtained decision boundaries that are straight along certain axes. If one wants to have a decision boundary that is straight along an arbitrary direction, the input space can be rotated at first to make the desired direction be aligned with certain axis, and a decision boundary that is straight along this axis can be produced like before and then the input space is rotated back.  

\begin{figure}[t]
	\centering
	\subfigure[Rank-deficient $\mathrm{W} \ (\mathrm{W} \in \mathbb{R}^{2 \times 2}, \ \mathrm{W}_{\cdot, 1}=0)$ maps 2D input space into a line, and then the curve $\sigma(\mathrm{W}\mathbf{x}+\mathbf{b})$ intersects the line with normal $ \mathbf{a} $ at a point whose pre-image in input space is a linear decision boundary.]{
		\label{} 
		\includegraphics[width=12cm]{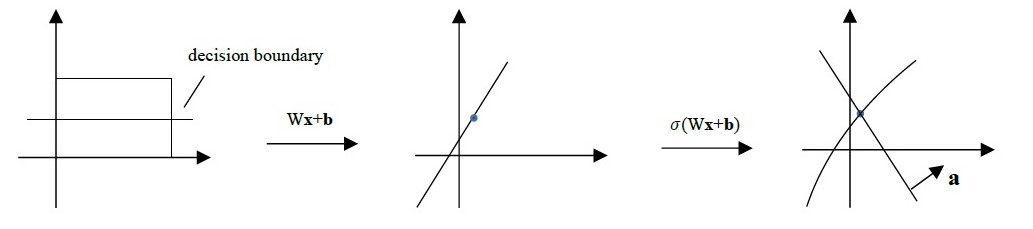}}
	\vspace{0.05cm}
	\subfigure[$\mathrm{W}\mathbf{x}+\mathbf{b}  \ (\mathrm{W} \in \mathbb{R}^{3 \times 2}, \ \mathrm{W}_{21}=\mathrm{W}_{31}=0)$ maps 2D input space into a plane $p$ in 3D, then the curved surface $\sigma(p)$ intersects the plane $q$ with normal $ \mathbf{a}=(0,1,1)^{T}$ at a line (both planes $p$ and $q$ are parallel to $x$ axis) whose pre-image in input space is a linear decision boundary.]{
		\label{} 
		\includegraphics[width=12cm]{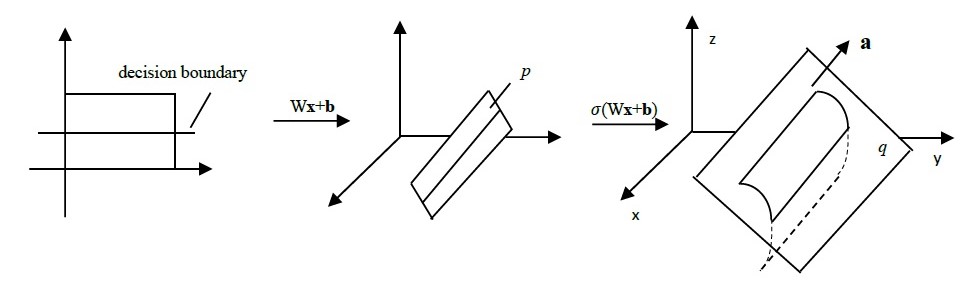}}
	\vspace{0.05cm}
	\subfigure[Rank-deficient $\mathrm{W} \ (\mathrm{W} \in \mathbb{R}^{3 \times 3}, \ \mathrm{W}_{\cdot, 1}=0)$ maps 3D input space into a plane $p,$ and its intersection with $\sigma^{-1}(q) \ (q$ is the plane with normal $ \mathbf{a} $ ) is a curve whose pre-image in 3D input space is a developable decision boundary.]{
		\label{} 
		\includegraphics[width=12cm]{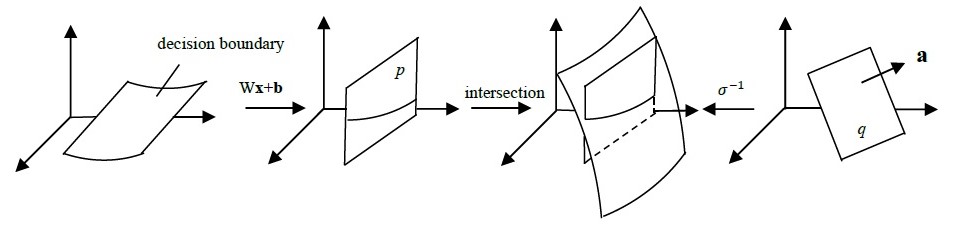}}
	\caption{Illustrations of how flat or developable decision boundaries are generated.}
	\label{} 
\end{figure}

\section{Some Topological Properties of Decision Boundaries}\label{section4}
In this section, based on the Gauss-Bonnet-Chern theorem for surfaces and higher-dimensional manifolds, we describe how to obtain the Euler characteristic of a compact decision boundary. 

\subsection{3D input space}\label{section4.1}
In this subsection, given a compact decision surface in 3D input space, we discuss how to compute its topological property in terms of Euler characteristic. The result is presented in  the following theorem, which is a direct consequence of the Gauss-Bonnet theorem for 2D surfaces.

\begin{thm}\label{theorm4.3}
	In 3D input space, for a compact orientable decision boundary surface $S$,  its Euler characteristic is $\chi(S)=  \frac{1}{2 \pi} \iint_{S} K \mathrm{d} \sigma $, where $K$ is the Gaussian curvature given in \eqref{eq4}, $\mathrm{d} \sigma$ is the area element.
	
\end{thm}

Example of compact surface and numerical computation of the integral $ \iint_{S} K \mathrm{d} \sigma $ will be presented in subsection \ref{section4.3}.

\subsection{Higher-dimensional input space}\label{section4.2}
For higher-dimensional input space, the topological structure of a compact orientable decision boundary can be characterized by its Euler characteristic. The following theorem shows that the Euler characteristic of a DNN's decision boundary can be obtained by an integral of local curvature, which is a direct application of the Gauss-Bonnet-Chern theorem for higher-dimensional manifolds.
\begin{thm}\label{theorm4.5}
	For a $2n$-dimensional compact orientable decision boundary manifold $M,$ its Euler characteristic $\chi(M)$ can be obtained by
	\begin{equation}\label{eq7}
	\chi(M)= \int_{M} e(M) ,
	\end{equation}
	where 
	\begin{equation}\label{eq8}
	\quad e=\frac{1}{2^{n}(2 \pi)^{n} n !} \sum_{a_{1}, a_{2}, \cdots a_{2 n}} \varepsilon_{a_{1}, a_{2}, \cdots a_{2 n}} \Omega^{a_{1} a_{2}} \wedge \cdots \wedge \Omega^{a_{2 n-1} a_{2 n}},
	\end{equation}
	and
	$\Omega^{a b}=\Sigma_{c, d} g^{a c} g^{b d} \Omega_{c d}$
	,\ $\varepsilon_{a_{1}, a_{2}, \cdots a_{2 n}}=\sqrt{\operatorname{det}(g)} \delta_{a_{1} \cdots a_{2 n}}^{1 \cdots 2 n} $.  The
	expressions of metric tensor $ g $ and curvature 2-form $\Omega_{a b}$ in terms of network weights will be given in subsection \ref{section7.3}.
\end{thm}
\textbf{Remark.} For high-dimensional input space, computing the integral in \eqref{eq7} faces two difficulties: the exponentially increasing number of terms in $e,$ and the high dimensions of integral. Efficient computation of this integral using techniques such as sampling will be one of our future work.

%Proof of Theorem \ref{theorm4.5}
%
%By  high-dimensional Gauss-Bonnet-Chern theorem, there is $\chi(M)=\int_{M} e(M)$. 
%Using Stokes' theorem for manifold: $\int_{M} e(M)=\int_{\Omega} \mathrm{d} e,$ where $\Omega$ is the volume enclosed by
%$M$, \eqref{eq7} is obtained immediately.

\subsection{Experiments on Topological Properties}\label{section4.3}

\subsubsection{3D input space}
In this subsection, we conduct experiments to show that in 3D input space, for sphere-like (genus zero) decision boundaries obtained by neural network training, the integral $ \iint_{S} K \mathrm{d} \sigma $ in Theorem \ref{theorm4.3} is indeed $ 2 \pi \chi(S)$, where $ \chi(S)=2 $ is the Euler characteristic of sphere-like surfaces, thus demonstrating the correctness of Theorem \ref{theorm4.3} for genus zero decision boundaries. 

\begin{figure}[t]
	\centering
	\subfigure[.]{
		\label{} 
		\includegraphics[width=7cm]{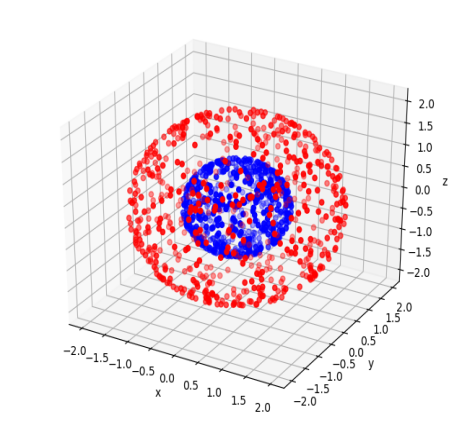}}
	\vspace{0.05cm}
	\subfigure[.]{
		\label{} 
		\includegraphics[width=4.5cm]{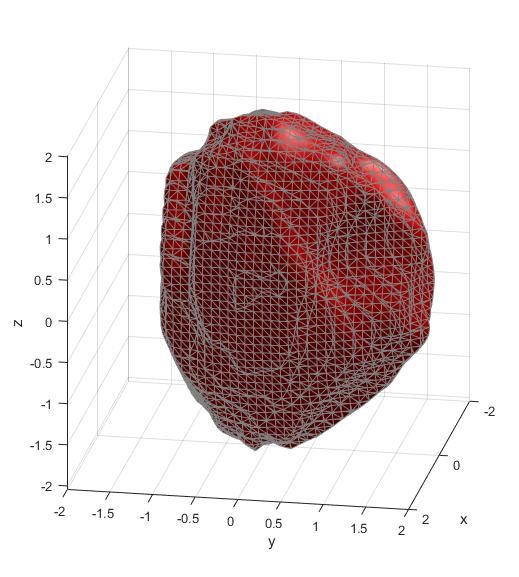}}
	\caption{ (a). a synthetic dataset for binary classification. (b). the decision boundary produced by a one-hidden-layer neural network.}
	\label{fig2} 
\end{figure}

We generate a synthetic dataset, shown in fig. \ref{fig2}(a), as follows for binary classification. For the first class, we draw 600 3D data points from the 3D Gaussian distribution with zero mean vector and identity covariance matrix. Each data point is then normalized to have unit vector length, i.e., these points are located on a unit sphere after normalization. Data points of the second class are obtained by scaling the data vectors of the first class such that they lie on the sphere with a radius of 2.

A one-hidden-layer neural network is trained to classify the data points. In the hidden layer, there are 40 hidden neurons with the tanh activation function. 
%A sigmoid activation is appended to the output neuron to squash the output into $ [0 \ 1] $. 
Network weights are initialized by a Gaussian distribution with zero mean and a variance of 0.01, and each bias is initialized to zero. Gradient descent is used to train the network with cross-entropy loss, and the learning rate is set to 0.5. The training stops after $ 5 \times 10^5 $ iterations, obtaining a training accuracy of $ 98.9\% $.

We now describe how to compute the integral $ \iint_{S} K \mathrm{d} \sigma $, where $K$ is the Gaussian curvature given in \eqref{eq4}, $\mathrm{d} \sigma$ is the surface element. We partition the decision boundary surface into a number of small patches, then according to the definition of surface integral, $ \iint_{S} K \mathrm{d} \sigma $ can be approximated by $ \sum_{i=0}^{n} K(\xi_i,\eta_i,\zeta_i) \Delta S_i$, where $ (\xi_i,\eta_i,\zeta_i) $ is any point in the $ i $th patch, $ \Delta S_i $ is its area. When the patches are sufficiently small, the integral $ \iint_{S} K \mathrm{d} \sigma $ can be obtained as such with high accuracy. Matlab is used to implement the computation. We use levelset of $ f(x,y,z)=0 $ to compute the decision boundary explicitly. We first use the meshgrid() function in Matlab to sample a 3D grid of points evenly spaced in the region of interest that contains all input data points, and the distance between two neighboring grid points is given by a parameter $ \lambda $. Then, the values of $ f(x,y,z) $ at all grid points are computed, and the isosurface() function in Matlab is utilized to find the decision boundary, defined as the levelset of $ f(x,y,z)=0 $, from the values at grid points. The isosurface() function returns the decision boundary in the form of faces and vertices of isosurface, and we then use the patch() function of Matlab to visualize the decision boundary through plotting filled polygonal faces. Fig. \ref{fig2}(b) shows the obtained decision boundary. Based on the faces and vertices of isosurface, we calculate the term $  K(\xi_i,\eta_i,\zeta_i) $ for each patch on the decision boundary using \eqref{eq4}. Each patch generated by isosurface() is a triangle, thus the area of each patch $ \Delta S_i $ can be computed by $ 0.5|\overrightarrow{ab} \times \overrightarrow{ac}| $, where $ a,b,c $ are the three vertices of a triangle. Finally,  integral $ \iint_{S} K \mathrm{d} \sigma $ is approximated by $ \sum_{i=0}^{n} K(\xi_i,\eta_i,\zeta_i) \Delta S_i$. 

In our experimental results, for the obtained genus zero (sphere-like) decision boundary, we have $ \iint_{S} K \mathrm{d} \sigma = 12.65$  when $ \lambda=0.02 $, which is very close to $ 4 \pi = 12.57$ and apparently different from $ \iint_{S} K \mathrm{d} \sigma$ for surfaces with other Euler characteristics ($ 2 \pi \chi(S)=  0, -4 \pi \ \text{etc}$), demonstrating the correctness of Theorem \ref{theorm4.3} for genus zero decision boundaries. The accuracy of evaluating $ \iint_{S} K \mathrm{d} \sigma$ can be increased with smaller $ \lambda$.

\section{Proofs}\label{section7}

\subsection{Proof of Lemma \ref{lemma3.1}}\label{section7.1}
The curvature of a planar curve can be found in some standard calculus textbooks (e.g. \cite{Curvature2d}) as follows,
\begin{equation}\label{eq9}
k=|\mathrm{y}^{\prime \prime}| \cdot\left(1+y^{\prime 2}\right)^{-\frac{3}{2}} ,
\end{equation}
where $y^{\prime}=\frac{\mathrm{d} y}{\mathrm{d} x}$ and $y^{\prime \prime}=\frac{\mathrm{d^2} y}{\mathrm{d} x^2} $. We then need to compute $y^{\prime}$ and $y^{\prime \prime}$ involved in \eqref{eq9}. Given the equation of decision boundary $f(x, y)=0$ which implicitly defines a curve $y=y(x),$ its derivative with respect to $x$ leads to
$$
\frac{\partial f}{\partial x}+\frac{\partial f}{\partial y} \cdot \frac{\mathrm{d} y}{\mathrm{d} x}=0 .
$$
So $y^{\prime}=-\frac{f_x}{f_y}$ and consequently 
\begin{equation}\label{eq41}
\begin{aligned}
y^{\prime \prime}&=-\frac{(f_{x x}+f_{x y} y^{\prime} ) f_{y}-f_{x}\left(f_{y x}+f_{y y} \cdot y \prime\right)}{{f_y}^{2}} \\  &=-\frac{f_{x x} f_{y}-2 f_{x} f_{y x}+\frac{{f_x}^{2}}{f_{y}} f_{y y}}{{f_y}^{2}}.
\end{aligned}
\end{equation}
Substituting them into \eqref{eq9}, we get \eqref{eq2}. 
%For convex decision boundary curves, $y^{\prime \prime}>0 .$ It then follows from \eqref{eq9} that $k_{r}>0$, which is consistent with the conclusion of differential geometry for convex curves.

\subsection{Proof of Lemma \ref{lemma3.2}}\label{section7.2}
The equation $f(x, y, z)=0$ defines a decision boundary surface $z=z(x, y)$ implicitly. The Gaussian curvature $K$ at a point $\mathbf{r}=(x, y, z(x, y))^{T}$ on the surface is given by \cite{Curvature3d}
\begin{equation}\label{eq40}
K=\frac{z_{xx}z_{yy} - z_{xy}^2}{(1+z_x^2++z_y^2)^2}
\end{equation}
From $f(x, y, z)=0,$ we have $\frac{\partial f}{\partial \mathrm{x}}+\frac{\partial f}{\partial z} \cdot \frac{\partial z}{\partial \mathrm{x}}=0,$ thus $z_x= \frac{\partial z}{\partial x}=-\frac{f_{x}}{f_{z}}$ and similarly $z_y= -\frac{f_{y}}{f_{z}}$. Similar to \eqref{eq41}, we can obtain $ z_{xx}, z_{yy} $ and $ z_{xy} $. Substituting them into \eqref{eq40} and through direct calculation, we can obtain \eqref{eq4}.

\subsection{Curvatures for higher-dimensional input space}\label{section7.3}

For high-dimensional input space, we need to compute the curvature tensor and curvature 2-form, which in turn depend on the metric tensor $(g_{i j})$ and Riemannian connection $\Gamma_{i k}^{m}$.

In $ d $-dimensional input space, the decision boundary is generally a $ (d-1) $-dimensional manifold. We use $x_{i} \ (i=1,2, \ldots, d-1)$ to parameterize the decision boundary, thus a point on the decision boundary is locally represented as $\mathbf{r}=\left(x_{1}, x_{2}, \cdots, x_{d-1}, x_{d}\right)^{T},$ where $x_{d}=x_{d}\left(x_{1}, x_{2}, \cdots, x_{d-1}\right)$ is implicitly
defined by the equation $f\left(x_{1}, x_{2}, \cdots, x_{d-1}, x_{d}\right)=0$. The metric tensor $g_{i j}$ is defined by $\mathrm{d} s^{2}=\sum_{i, j=1}^{d-1} g_{i j} \mathrm{d} x_{i} \mathrm{d} x_{j}$. Therefore, by
$$
\begin{aligned}
\mathrm{d} s^{2}&=\mathrm{d} \mathbf{r} \cdot \mathrm{d} \mathbf{r}=\mathrm{d} x_{1}^{2}+\mathrm{d} x_{2}^{2}+\cdots+\mathrm{d} x_{d}^{2} \\
&=\mathrm{d} x_{1}^{2}+\mathrm{d} x_{2}^{2}+\cdots+\mathrm{d} x_{d-1}^{2}+\left(\sum_{i=1}^{d-1} \frac{\partial x_{d}}{\partial x_{i}} \mathrm{d} x_{i}\right)^{2},
\end{aligned}
$$
and by $\frac{\partial x_{d}}{\partial x_{i}}=-f_{i} / f_{d},$ there is $\mathrm{d} s^{2}=\sum_{i=1}^{d-1} \mathrm{d} x_{i}^{2}+\sum_{i, j=1}^{d-1} \frac{f_{i} f_{j}}{f_{d}^{2}} \mathrm{d} x_{i} \mathrm{d} x_{j} .$ We get the components of
$(g_{i j})$ as follows, 
\begin{equation}\label{eq10}
%\begin{array}{c}
\begin{aligned}
g_{i i}&=1+f_{i}^{2} / f_{d}^{2}, \quad i=1,2, \cdots, d-1,  \\ 
g_{i j}&=f_{i} f_{j} / f_{d}^{2}, \quad i, j=1,2, \cdots, d-1, \text { and } i \neq j.
%\end{array}
\end{aligned}
\end{equation}

Having metric tensor $g=\left(g_{i j}\right),$ we can compute the Riemannian connection defined as
$$
\Gamma_{i k}^{m} \ := \frac{1}{2} \Sigma_{j} g^{m j}\left(\partial_{k} g_{i j}+\partial_{i} g_{k j}-\partial_{j} g_{i k}\right),
$$
where $g^{m j}$ is the element of $g^{-1}$. The computation is very lengthy and tedious, hence we only show the key techniques to save space. For example, $\partial_{k} g_{i j}$ is computed as follows,
\begin{equation}\label{eq11}
\begin{aligned}
\partial_{k} g_{i j} &=\frac{1}{f_{d}^{4}}  \left[\left(f_{i k}+f_{i d} \cdot \frac{\partial x_{d}}{\partial x_{k}}\right) f_{j}+f_{i}\left(f_{j k}+f_{j d} \frac{\partial x_{d}}{\partial x_{k}}\right)\right] f_{d}^{2}  -\frac{f_{i} f_{j}}{f_{d}^{4}}  \cdot\left[2 f_{d}\left(f_{d k}+f_{d d} \frac{\partial x_{d}}{\partial x_{k}} \right) \right]  \\
&=  [   f_{j} f_{d}^{2} f_{i k}+f_{i} f_{d}^{2} f_{j k}-f_{j} f_{k} f_{d} f_{i d}-f_{i} f_{k} f_{d} f_{j d} -2 f_{i} f_{j} f_{d} f_{d k}+2 f_{i} f_{j} f_{k} f_{d d} ] / f_{d}^{4}, \ \ (i \neq j), \\  
\partial_{k} g_{i i} &=\frac{2 f_{i} f_{d}^{2} f_{i k}-2 f_{i} f_{k} f_{d} f_{i d}+2 f_{i}^{2} f_{k} f_{d d}-2 f_{i}^{2} f_{d} f_{d k}}{f_{d}^{4}}.
\end{aligned}
\end{equation}
The Riemann curvature tensor is defined by
\begin{equation}\label{eq12}
R_{b i k}^{a}=\Gamma_{k b, i}^{a}-\Gamma_{i b, k}^{a}+\sum_{c} \Gamma_{i c}^{a} \Gamma_{k b}^{c}-\sum_{c} \Gamma_{k c}^{a} \Gamma_{i b}^{c}.
\end{equation}
The key computation here is $\Gamma_{k b, i}^{a} := \frac{\partial \Gamma_{k b}^{a}}{\partial x_{i}},$ which involves the computation of $\frac{\partial g^{m j}}{\partial x_{i}} .$ By $\Sigma_{j} g^{m j} g_{j n}=\delta_{n}^{m},$ differentiating it gives
$$
\Sigma_{j}\left(\frac{\partial g^{m j}}{\partial x_{i}} g_{j n}+g^{m j} \frac{\partial g_{j n}}{\partial x_{i}}\right)=0.
$$
Rewrite in matrix form,
$$
\frac{\partial g^{-1}}{\partial x_{i}} g=-g^{-1} \frac{\partial g}{\partial x_{i}}.
$$
Therefore, we have
\begin{equation}\label{eq13}
\frac{\partial g^{m j}}{\partial x_{i}}=-\sum_{k, n} g^{m k} \frac{\partial g_{k n}}{\partial x_{i}} g^{n j}.
\end{equation}

Finally, the curvature 2-form is
\begin{equation}\label{eq14}
\Omega_{a b} := \frac{1}{2} \sum_{i, k} R_{a b i k} \mathrm{d} x^{i} \wedge \mathrm{d} x^{k}=\frac{1}{2} \sum_{i, k, n} g_{a n} R_{b i k}^{n} \mathrm{d} x^{\mathrm{i}} \wedge \mathrm{d} x^{k}.
\end{equation}
$\Omega_{a b}$ is used in Gauss-Bonnet-Chern theorem to compute Euler characteristics of compact manifolds.

\subsection{Proof of Theorem \ref{theorm6.1}}
For a one-hidden-layer network $f(\mathbf{x})=\mathbf{a}^{\mathrm{T}} \sigma(\mathrm{W}\mathbf{x}+\mathbf{b})+c,$ its $1st$-order and $2nd$-order partial derivatives have been given in \eqref{eq3}. By \eqref{eq2}, we can obtain the expression of $k$ in terms of network weights. Here we only concern the following part of $k$ that will cause $k=0$.
\begin{equation}\label{eq30}
%\begin{array}{l}
\begin{aligned}
&f_{x x} f_{y}^{2}  -2 f_{x} f_{y} f_{x y}+f_{x}^{2} f_{y y}  \\ 
&=\sum_{i, j, k} [ \mathrm{a}_{i} \sigma_{i}^{\prime \prime} \mathrm{W}_{i 1}^{2} \cdot \mathrm{a}_{j} \sigma_{j}^{\prime} \mathrm{W}_{j 2} \cdot \mathrm{a}_{k} \sigma_{k}^{\prime} \mathrm{W}_{k 2} -2 \mathrm{a}_{i} \sigma_{i}^{\prime \prime} \mathrm{W}_{i 1} \mathrm{W}_{i 2} \cdot \mathrm{a}_{j} \sigma_{j}^{\prime} \mathrm{W}_{j 1} \cdot \mathrm{a}_{k} \sigma_{k}^{\prime} \mathrm{W}_{k 2} +\mathrm{a}_{i} \sigma_{i}^{\prime \prime} \mathrm{W}_{i 2}^{2} \cdot \mathrm{a}_{j} \sigma_{j}^{\prime} \mathrm{W}_{j 1} \mathrm{a}_{k} \sigma_{k}^{\prime} \mathrm{W}_{k 1} ]  \\ 
&=\sum_{i, j, k} \sigma_{i}^{\prime \prime} \sigma_{j}^{\prime} \sigma_{k}^{\prime} \cdot \mathrm{a}_{i} \mathrm{a}_{j} \mathrm{a}_{k} \cdot [ \mathrm{W}_{i 1} \mathrm{W}_{k 2}\left(\mathrm{W}_{i 1} \mathrm{W}_{j 2}-\mathrm{W}_{i 2} \mathrm{W}_{j 1}\right)  +\mathrm{W}_{i 2} \mathrm{W}_{j 1}\left(\mathrm{W}_{i 2} \mathrm{W}_{k 1}-\mathrm{W}_{i 1} \mathrm{W}_{k 2}\right) ] \\
&=0 .
\end{aligned}
%\end{array}
\end{equation}
For general activation functions, $\sigma^{\prime \prime}(y)$ and $\sigma^{\prime}(y)$ cannot equal zero everywhere. Therefore by \eqref{eq30}, $k(\mathbf{x})=0$ at any point $\mathbf{x}$ on the decision boundary requires
\begin{equation}\label{eq31}
%\begin{array}{c}
\begin{aligned}
\mathrm{a}_{i} \mathrm{a}_{j} \mathrm{a}_{k} [  \mathrm{W}_{i 1} \mathrm{W}_{k 2}\left(\mathrm{W}_{i 1} \mathrm{W}_{j 2}-\mathrm{W}_{i 2} \mathrm{W}_{j 1}\right) 
+  \mathrm{W}_{i 2} \mathrm{W}_{j 1}\left(\mathrm{W}_{i 2} \mathrm{W}_{k 1}-\mathrm{W}_{i 1} \mathrm{W}_{k 2}\right) ]=0,   \quad\left( \forall i, j, k \in\left[d_{1}\right]\right) .
\end{aligned}
%\end{array}
\end{equation}
There are serval possible non-trivial (i.e., $\mathbf{a} \neq 0$ and $\mathrm{W} \neq 0$ ) cases making \eqref{eq31} satisfied, which constitute the sufficient conditions for $k(\mathbf{x})=0$.
\\ 

\textbf{a)} $\mathrm{a}_{i} \mathrm{W}_{i 1}=0$ or $\mathrm{a}_{i} \mathrm{W}_{i 2}=0 \ (\forall i \in [d_1]) $. So, if $\mathrm{a}_{i} \neq 0,$ there is $\mathrm{W}_{i 1}=0$ or $\mathrm{W}_{i 2}=0 .$ Let $I=\left\{i \in \mathbb{R}^{d_{1}} | \mathrm{a}_{i} \neq 0\right\},$ we have (assume $\mathrm{a}_{i} \mathrm{W}_{i 1}=0$)
$$
f(\mathbf{x})=\sum_{i \in I} a_{i} \sigma\left(\mathrm{W}_{i 2} y+b_{i}\right)+c=0
$$
due to $\mathrm{W}_{i 1}=0 \ (i \in I) .$ This equation only involves variable $y .$ As a result, its solution $y^{*}$ determines a linear decision boundary that is parallel to $x$ axis.
\\

\textbf{b)} $\mathrm{a}_{i} \mathrm{a}_{j}\left(\mathrm{W}_{i 1} \mathrm{W}_{j 2}-\mathrm{W}_{i 2} \mathrm{W}_{j 1}\right)=0 \ (\forall i, j \in [d_1])$. 
$\mathrm{W}_{i 1} \mathrm{W}_{j 2}-\mathrm{W}_{i 2} \mathrm{W}_{j 1}=0$ means the $i$th row and $j$th row of $\mathrm{W}$ are linear dependent, hence vectors in $  \left \{  \mathbf{w}_{i} | i \in I  \right \}   $ are linear dependent.  Without loss of generality, suppose  $ \mathbf{w}_{1} \neq 0 $ and $ 1 \in I $, thus $\mathbf{w}_{i}=c_{i} \mathbf{w}_{1} \ \left(i \in [k], c_{i} \text { is a constant}\right) .$ Then, we have
$$
f(\mathbf{x})=\sum_{i \in I} \mathrm{a}_{i} \sigma\left(c_{i} \mathbf{w}_{1} \cdot \mathbf{x}+b_{i}\right)+c=0.
$$
$f(\mathbf{x})$ now only involves $\mathbf{w}_{1} \cdot \mathbf{x} $. The solution $\mathbf{w}_{1} \cdot \mathbf{x}=d^{*}$ to $ f(\mathbf{x})=0 $
produces a linear decision boundary in input space whose normal is $\mathbf{w}_{1}$.

\subsection{Proof of Theorem \ref{theorm6.3}}
Given the derivatives for one-hidden-layer networks in \eqref{eq3}, the numerator of $K$ in \eqref{eq4} can be expressed as
\begin{equation}\label{eq34}
\begin{aligned}
%\begin{array}{l}
&\left(2 f_{x z} f_{x} f_{z}-f_{x x} f_{z}^{2}-f_{z z} f_{x}^{2}\right)\left(2 f_{y z} f_{y} f_{z}-f_{y y} f_{z}^{2}-f_{z z} f_{y}^{2}\right)  -\left(f_{y z} f_{x} f_{z}+f_{x z} f_{y} f_{z}-f_{z z} f_{x} f_{y}-f_{x y} f_{z}^{2}\right)^{2} \\
&= \sum_{i, j, k, l, m, n} \mathrm{a}_{i} \mathrm{a}_{j} \mathrm{a}_{k} \mathrm{a}_{l} \mathrm{a}_{m} \mathrm{a}_{n} \sigma_{i}^{\prime \prime} \sigma_{j}^{\prime} \sigma_{k}^{\prime} \sigma_{l}^{\prime \prime} \sigma_{m}^{\prime} \sigma_{n}^{\prime} \\ 
& \quad \quad \quad\quad  \left [  \mathrm{W}_{i 1} \mathrm{W}_{k 3} \left( \mathrm{W}_{i 3} \mathrm{W}_{j 1}-\mathrm{W}_{i 1} \mathrm{W}_{j 3}\right) + \mathrm{W}_{i 3} \mathrm{W}_{j 1} \left( \mathrm{W}_{i 1} \mathrm{W}_{k 3}-\mathrm{W}_{i 3} \mathrm{W}_{k 1} \right) \right ] \cdot  \\ 
&  \quad \quad \quad\quad [ \mathrm{W}_{l 2} \mathrm{W}_{n 3}\left(\mathrm{W}_{l 3} \mathrm{W}_{m 2}-\mathrm{W}_{l 2} \mathrm{W}_{m 3}\right) + \mathrm{W}_{l 3} \mathrm{W}_{m 2}\left(\mathrm{W}_{l 2} \mathrm{W}_{n 3}-\mathrm{W}_{l 3} \mathrm{W}_{n 2}\right) ] \\
& \ \  -\sum_{i, j, k, l, m, n} \mathrm{a}_{i} \mathrm{a}_{j} \mathrm{a}_{k} \mathrm{a}_{l} \mathrm{a}_{m} \mathrm{a}_{n} \sigma_{i}^{\prime \prime} \sigma_{j}^{\prime} \sigma_{k}^{\prime} \sigma_{l}^{\prime \prime} \sigma_{m}^{\prime} \sigma_{n}^{\prime} \\
& \quad \quad \quad\quad [ \mathrm{W}_{i 3} \mathrm{W}_{j 1}\left(\mathrm{W}_{i 2} \mathrm{W}_{k 3}-\mathrm{W}_{i 3} \mathrm{W}_{k 2}\right)+\mathrm{W}_{i 1} \mathrm{W}_{k 3}\left(\mathrm{W}_{i 3} \mathrm{W}_{j 2}-\mathrm{W}_{i 2} \mathrm{W}_{j 3}\right) ] \cdot \\ 
&  \quad \quad \quad\quad  [ \mathrm{W}_{l 3} \mathrm{W}_{m 1}\left(\mathrm{W}_{l 2} \mathrm{W}_{n 3}-\mathrm{W}_{l 3} \mathrm{W}_{n 2}\right)+\mathrm{W}_{l 1} \mathrm{W}_{n 3}\left(\mathrm{W}_{l 3} \mathrm{W}_{m 2}-\mathrm{W}_{l 2} \mathrm{W}_{m 3}\right)  ]  \\
& = 0 . 
%\end{array}
\end{aligned}
\end{equation}
The following conditions are sufficient to make \eqref{eq34} satisfied: $\forall i, j, k \in\left[d_{1}\right]$, 
\begin{equation}\label{eq35}
%\begin{array}{l}
\begin{aligned}
& \quad \mathrm{a}_{i} \mathrm{a}_{j} \mathrm{a}_{k} [ \mathrm{W}_{i 1} \mathrm{W}_{k 3}\left(\mathrm{W}_{i 3} \mathrm{W}_{j 1}-\mathrm{W}_{i 1} \mathrm{W}_{j 3}\right)  +\mathrm{W}_{i 3} \mathrm{W}_{j 1}\left(\mathrm{W}_{i 1} \mathrm{W}_{k 3}-\mathrm{W}_{i 3} \mathrm{W}_{k 1}\right) ]=0 \\
\text{or} &\quad \mathrm{a}_{i} \mathrm{a}_{j} \mathrm{a}_{k} [ \mathrm{W}_{i 2} \mathrm{W}_{k 3}\left(\mathrm{W}_{i 3} \mathrm{W}_{j 2}-\mathrm{W}_{i 2} \mathrm{W}_{j 3}\right)  +\mathrm{W}_{i 3} \mathrm{W}_{j 2}\left(\mathrm{W}_{i 2} \mathrm{W}_{k 3}-\mathrm{W}_{i 3} \mathrm{W}_{k 2}\right) ]=0, \\
\text{and} &\ \ \mathrm{a}_{i} a_{j} \mathrm{a}_{k} [ \mathrm{W}_{i 3} \mathrm{W}_{j 1}\left(\mathrm{W}_{i 2} \mathrm{W}_{k 3}-\mathrm{W}_{i 3} \mathrm{W}_{k 2}\right)  +\mathrm{W}_{i 1} \mathrm{W}_{k 3}\left(\mathrm{W}_{i 3} \mathrm{W}_{j 2}-\mathrm{W}_{i 2} \mathrm{W}_{j 3}\right) ]=0, 
\end{aligned}
%\end{array}
\end{equation}
which further lead to the following sufficient conditions for $ K=0 $.
\\

\textbf{a)} $\mathrm{a}_{i} \mathrm{W}_{i 1}=0$ or $\mathrm{a}_{i} \mathrm{W}_{i 2}=0$ or $\mathrm{a}_{i} \mathrm{W}_{i 3}=0 \ \left(\forall i \in\left[d_{1}\right]\right)$. As an example, $\mathrm{a}_{i} \mathrm{W}_{i 1}=0$ means $\mathrm{W}_{i 1}=0 \ (i \in I) .$ Thus, on the decision boundary,
$$
f(\mathbf{x})=\sum_{i \in I} \mathrm{a}_{i} \sigma\left(\mathrm{W}_{i 2} y+\mathrm{W}_{i 3} z+\mathrm{b}_{i}\right)+c=0.
$$
This is an equation of $y$ and $z,$ giving an implicit curve $z=z(y) .$ The pre-image of this curve in 3D input space is a developable surface that is straight along $x$ direction.
\\

\textbf{b)} $\mathrm{a}_{i} \mathrm{a}_{j}\left(\mathrm{W}_{i 3} \mathrm{W}_{j 2}-\mathrm{W}_{i 2} \mathrm{W}_{j 3}\right)=0 \ \left(\forall i, j \in \left[d_{1}\right]\right) .$ This case also generates developable decision boundaries. The reason is as follows. We have
$$
f(\mathbf{x})=\sum_{i \in I} \mathrm{a}_{i} \sigma\left(\mathrm{W}_{i 1} x+\mathrm{W}_{i 2} y+\mathrm{W}_{i 3} z+\mathrm{b}_{i}\right)+c=0.
$$
Choose an index $k \in I$ for which $\mathbf{w}_{k} \neq 0,$ then $\mathrm{W}_{i 3} \mathrm{W}_{j 2}-\mathrm{W}_{i 2} \mathrm{W}_{j 3}=0 \ (\forall i, j \in I)$ implies $\mathbf{w}_{j, 2 \sim 3}=$
$c_{j} \mathbf{w}_{k, 2 \sim 3} \ (j \in I)$, where $\mathbf{w}^T_{j, 2 \sim 3}$ denotes the vector composed of the 2nd and $3 \mathrm{rd}$ columns of $\mathbf{w}^T_{j}$, $ c_{j} $ is a constant. Define a new variable $y^{\prime}=\mathrm{W}_{k 2} y+\mathrm{W}_{k 3} z,$ then
$$
f(\mathbf{x})=\sum_{i \in I} \mathrm{a}_{i} \sigma\left(\mathrm{W}_{i 1} x+c_{i} y^{\prime}+\mathrm{b}_{i}\right)+c=0.
$$
This is an equation of $x$ and $y^{\prime}$ whose solution is $y^{\prime}=y^{\prime}(x) .$ Therefore, for any value of $x,$ we have
$$
\mathrm{W}_{k 2} y+\mathrm{W}_{k 3} z=y^{\prime}(x),
$$
which specifies a line parallel to the $y-z$ plane with normal $\left(0, \mathrm{W}_{k 2}, \mathrm{W}_{k 3}\right)^{\top}$. As a result, the whole decision boundary is a developable surface.

\subsection{Proof of Theorem \ref{theorm6.4}}\label{section7.6}

The condition $\mathrm{a}_{i} \mathrm{W}_{i j}^{L} \mathrm{W}_{j k}^{L-1} \ldots \mathrm{W}_{m n}^{2}=0 \ (\forall i \in \mathbb{R}^{d_{L}}, j \in \mathbb{R}^{d_{L-1}}, k \in \mathbb{R}^{d_{L-2}}, \cdots, m \in \mathbb{R}^{d_{2}} )$ makes the term $\sigma\left(\mathbf{w}_{n}^{1} \cdot \mathbf{x}+b^{1}_n\right)$ vanish in $f(\mathbf{x})$,
thus $f(\mathbf{x})=0$ only involves $\mathbf{w}_{l}^{1} \cdot \mathbf{x} \ (\forall l \neq n)$. Linear dependency among $\mathbf{w}_{l}^{1} \ (\forall l \neq n)$ will give a linear decision boundary with normal $\mathbf{w}_{p}^{1}$ (any $p \neq n$ and $ \mathbf{w}_{p}^{1} \neq 0 $).

When $\mathrm{W}_{l q}^{1}=0 \ (\forall l \neq n)$, $x_{q}$ will disappear in equation $f(\mathbf{x})=0,$ and the solution to $f\left(x_{1}, x_{2}, \cdots x_{q-1}, x_{q+1} \cdots x_{d}\right)=0$ is an implicit hypersurface which is straight along $x_{q} .$ This conclusion still holds when the cardinality of $q$ is greater than $1,$ i.e., the decision boundary is straight along multiple axes.

\section{Conclusion and future work}

We have explored some geometrical and topological properties of decision boundaries produced by DNNs using knowledge of differential geometry. Based on the curvatures of decision boundaries in terms of network parameters, we give sufficient conditions on neural network parameters in order to produce flat or developable decision boundaries.  Based on the  Gauss-Bonnet-Chern theorem in differential geometry, we then proposed a method to obtain the Euler characteristics of compact decision boundaries using a particular integral of curvatures, and conducted experiments to compute the Euler characteristics of sphere-like decision boundaries generated by neural networks.

In future work, we plan to apply efficient numerical techniques to compute the integrals in section \ref{section4}  for topological properties of decision boundaries in high-dimensional input space. We are also interested in exploring the constraints on network architecture and parameters in order to produce decision boundaries with more complex geometrical properties, such as convexity or constant curvatures.

\bibliographystyle{plain}
\bibliography{GeometryToplogyDNNs}

\end{document}